\documentclass[letterpaper]{article} % DO NOT CHANGE THIS
\usepackage{aaai2026}  % DO NOT CHANGE THIS
\usepackage{times}  % DO NOT CHANGE THIS
\usepackage{helvet}  % DO NOT CHANGE THIS
\usepackage{courier}  % DO NOT CHANGE THIS
\usepackage[hyphens]{url}  % DO NOT CHANGE THIS
\usepackage{graphicx} % DO NOT CHANGE THIS
\usepackage{natbib}  % DO NOT CHANGE THIS AND DO NOT ADD ANY OPTIONS TO IT
\usepackage{caption} % DO NOT CHANGE THIS AND DO NOT ADD ANY OPTIONS TO IT
\usepackage{algorithm}

\usepackage{cite}
\usepackage{graphicx}
\usepackage{booktabs}
\usepackage{subfigure}
\usepackage{algorithm}
\usepackage{algpseudocode}
\usepackage{amsthm,amsmath,amssymb}
\usepackage{bm}
\usepackage{multirow}
\usepackage{bbding}
\usepackage{pifont}
\usepackage{wasysym}
\usepackage{tabularx}
\usepackage{makecell}

\usepackage{newfloat}
\usepackage{listings}
\DeclareCaptionStyle{ruled}{labelfont=normalfont,labelsep=colon,strut=off} % DO NOT CHANGE THIS
\floatstyle{ruled}
\newfloat{listing}{tb}{lst}{}
\floatname{listing}{Listing}
\title{Other Vehicle Trajectories Are Also Needed: A Driving World Model Unifies Ego-Other Vehicle Trajectories in Video Latent Space}
\author{
    Jian Zhu\thanks{Corresponding Author~(jianzhu823@gmail.com)} , Zhengyu Jia,  Tian Gao, Jiaxin Deng, Shidi Li, \\
Lang Zhang\thanks{Project Leader}, Fu Liu, Peng Jia, Xianpeng Lang\\
}
\affiliations{
    Li Auto Inc.\\
}

\usepackage{bibentry}
\begin{document}

\maketitle

\begin{figure*}[t]
   \centering
   \includegraphics [width=0.75\linewidth] {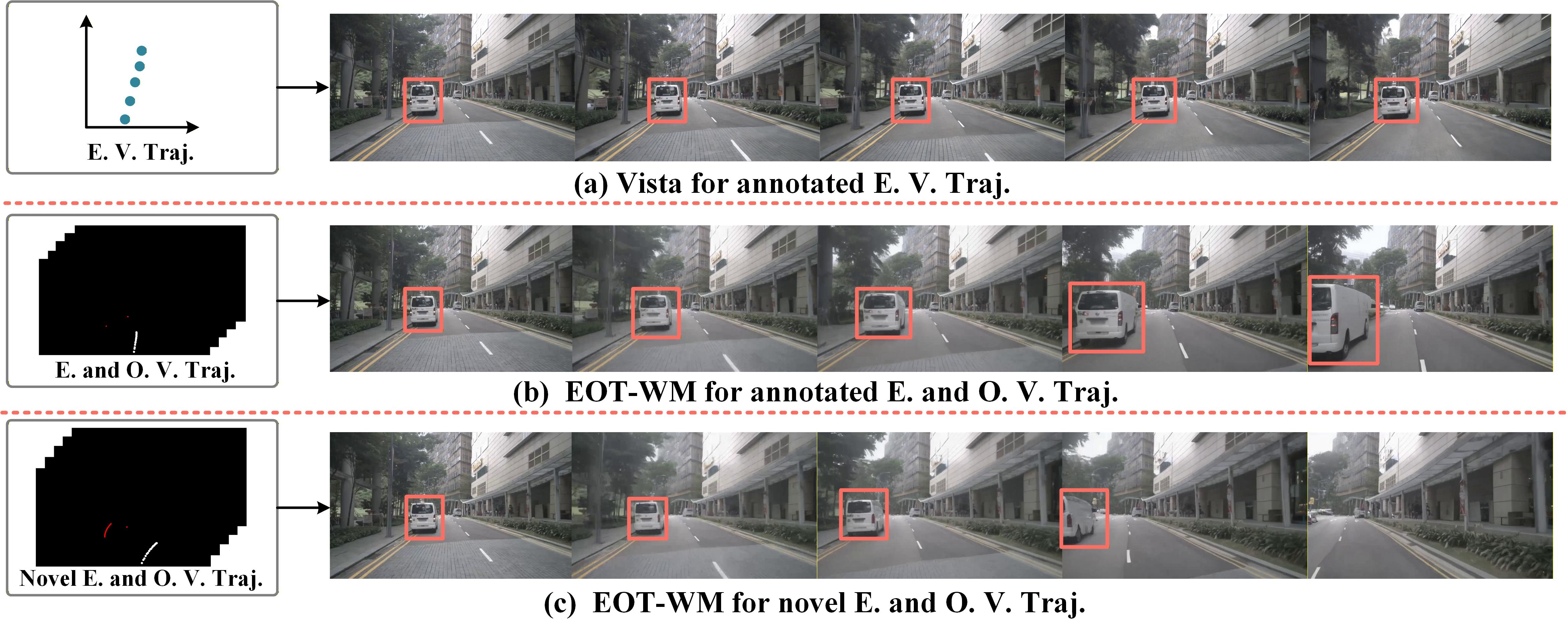}
   \caption{The proposed EOT-WM is capable of generating more realistic videos with controllable ego and other vehicle trajectories. These trajectories are represented in video space for EOT-WM instead of BEV space for previous works such as Vista. E. V. Traj. and O. V. Traj. denote ego and other vehicle trajectories, respectively. Novel trajectory means self-produced trajectory not included in the dataset.
   } \label{fig1}
\end{figure*}

\begin{abstract}
Advanced end-to-end autonomous driving systems predict other vehicles' motions and plan ego vehicle's trajectory. The world model that can foresee the outcome of the trajectory has been used to evaluate the autonomous driving system. However, existing world models predominantly emphasize the trajectory of the ego vehicle and leave other vehicles uncontrollable. This limitation hinders their ability to realistically simulate the interaction between the ego vehicle and the driving scenario.
In this paper, we propose a driving \textbf{W}orld \textbf{M}odel named EOT-WM, unifying \textbf{E}go-\textbf{O}ther vehicle \textbf{T}rajectories in videos for driving simulation. Specifically, it remains a challenge to match multiple trajectories in the BEV space with each vehicle in the video to control the video generation. We first project ego-other vehicle trajectories in the BEV space into the image coordinate for vehicle-trajectory match via pixel positions. Then, trajectory videos are encoded by the Spatial-Temporal Variational Auto Encoder to align with driving video latents spatially and temporally in the unified visual space. A trajectory-injected diffusion Transformer is further designed to denoise the noisy video latents for video generation with the guidance of ego-other vehicle trajectories. In addition, we propose a metric based on control latent similarity to evaluate the controllability of trajectories. Extensive experiments are conducted on the nuScenes dataset, and the proposed model outperforms the state-of-the-art method by 30\% in FID and 55\% in FVD. The model can also predict unseen driving scenes with self-produced trajectories.
\end{abstract}

% Uncomment the following to link to your code, datasets, an extended version or similar.
% You must keep this block between (not within) the abstract and the main body of the paper.
% \begin{links}
%     \link{Code}{https://aaai.org/example/code}
%     \link{Datasets}{https://aaai.org/example/datasets}
%     \link{Extended version}{https://aaai.org/example/extended-version}
% \end{links}

% Uncomment the following to link to your code, datasets, an extended version or similar.
% You must keep this block between (not within) the abstract and the main body of the paper.
% \begin{links}
%     \link{Code}{https://aaai.org/example/code}
%     \link{Datasets}{https://aaai.org/example/datasets}
%     \link{Extended version}{https://aaai.org/example/extended-version}
% \end{links}

\section{Introduction}
End-to-end autonomous driving~\cite{hu2023planning,tian2024drivevlm, li2025generative} has gained increasing attention recently, since the approach integrates all modules~(e.g. perception, decision, and planning modules) into a model optimized jointly to directly output planning results based on input multi-sensor data. Despite the promising performance of the end-to-end autonomous driving model, effectively handling out-of-distribution scenarios continues to be a significant challenge, particularly since such situations are often hazardous and costly to simulate. The world model~\cite{wang2023drivedreamer, wang2024driving, gao2024vista, hassan2025gem} can predict future driving scenes based on historical observations and future driving actions, which is a potential solution to evaluate the autonomous driving model and avoid catastrophic errors.

Some world models~\cite{gao2024vista, yang2024generalized} explore predicting future driving scenes with the trajectory, since it can reflect the driving action of the vehicle in the scene more precisely. Despite advanced end-to-end autonomous driving systems (e.g. VAD~\cite{jiang2023vad}) can predict other vehicles' motions and plan ego vehicle's trajectory, these world models primarily focus on the ego vehicle trajectory. They view the trajectory as a series of points in the bird’s eye view (BEV) space~\cite{hu2021fiery,li2023delving}, which are directly encoded as the condition for video generation. However, there are three main drawbacks in the approaches mentioned above. Firstly, they just consider the ego vehicle trajectory and leave other vehicles uncontrollable in the generated video. As a result, the model cannot realistically simulate the interaction between the ego vehicle and the driving scenario, and generate diverse novel scenes by changing other vehicle trajectories as well. Secondly, the distribution of the encoding based on the trajectory points in the BEV space is quite different from that of the video latents in visual modality without aligning their feature space. In addition, it is impractical to correspond multiple trajectories in the BEV space to numerous vehicles in visual modality, since the BEV space is mismatched with the video space.

To tackle above issue, a driving world model unifying ego-other vehicle
trajectories in videos named EOT-WM is proposed in this paper. As is shown in Fig.~\ref{fig1}, Vista~\cite{gao2024vista} only uses the ego vehicle trajectory, and the stationary other vehicle in the groundtruth video also moves forward in the generated video. The proposed EOT-WM can generate more realistic videos with controllable ego and other vehicle actions. In addition, the model can also generate video matched based on novel self-produced trajectories. Specifically, instead of representing the trajectory via points in the BEV space, we project these points into the image coordinate and plot the ego vehicle trajectory and the other vehicle trajectories in separate blank videos to generate trajectory videos for learning in unified visual modality. Then, we adopt Spatial-Temporal Variational Auto Encoder~(STVAE) to encode the scene video and trajectory video to achieve scene video latents and trajectory latents with sharing feature space. Moreover, the scene and trajectory latents achieved in this manner are aligned temporally and spatially to realize effective control. Finally, we design Trajectory-injected Diffusion Transformer~(TiDiT) to integrate motion guidances provided by trajectory latents into video latents for denoising the noisy video latents more precisely. As a result, the entire model can predict future frames based on the given initial frames with the control of text and trajectory. To evaluate the controllability of trajectories, we propose a metric based on control latent similarity, which compares the predicted trajectory latents with the groundtruth trajectory latents.

The main contributions of the paper can be summarized as follows.
\begin{itemize}
\item We firstly propose a driving world model with ego and other vehicle trajectories, more realistically simulating the interaction between the ego vehicle and the driving scenario and able to generate diverse novel scenes with alterable ego-other trajectories.
\item We propose to represent the trajectory as the video and encode the trajectory video via the driving video encoder to make each trajectory aligned with each vehicle in the unified visual space.
\item A trajectory-injected diffusion Transformer is designed to denoise the noisy video latents more precisely via the ego-other vehicle trajectories. A metric based on control latent similarity is further proposed to evaluate the controllability of trajectories.
% \item We conduct comprehensive experiments on the nuScenes dataset to evaluate the proposed EOT-WM, which outperforms the state-of-the-art method by 30\% in FID and 55\% in FVD.
\end{itemize}

\section{Related Work}

\subsection{Video Generation}

Video generation is capable of understanding the world and generating realistic video samples. Various kinds of models have been studied in the past, including VAE-based~\cite{villegas2019high, franceschi2020stochastic}, flow-based~\cite{dorkenwald2021stochastic,kumar2020videoflow}, GAN-based~\cite{brooks2022generating, yu2022generating} and auto-regressive models~\cite{ge2022long, weissenbornscaling}. Recently, diffusion models have achieved breakthroughs in image generation~\cite{nichol2022glide, rombach2022high}, and diffusion models are also introduced into video generation~\cite{blattmann2023align, guoanimatediff, ho2022video, yang2024cogvideox}. Most of these models~\cite{blattmann2023align, guoanimatediff, ho2022video, yang2024cogvideox} use the text as the condition to control video generation. To explore more condition forms for controlling video generation, image-to-video generative models~\cite{blattmann2023stable, chen2023videocrafter1} are studied since the image can provide more specific priors. However, above models~\cite{blattmann2023align, guoanimatediff, ho2022video, yang2024cogvideox, blattmann2023stable, chen2023videocrafter1} are incapable of predicting the future state, which are crucial in autonomous driving. In addition, camera motion~\cite{wang2024motionctrl} and object motion~\cite{wang2024motionctrl, zhang2024tora} are employed to realize more flexible video generation. Unfortunately, these models are not specifically designed to tackle complicated autonomous driving videos and cannot model vehicle actions well. In this paper, we propose EOT-WM that jointly uses texts and trajectories to predict future video frames based on given initial states.

\begin{figure*}[ht]
   \centering
   \includegraphics [width=0.65\linewidth] {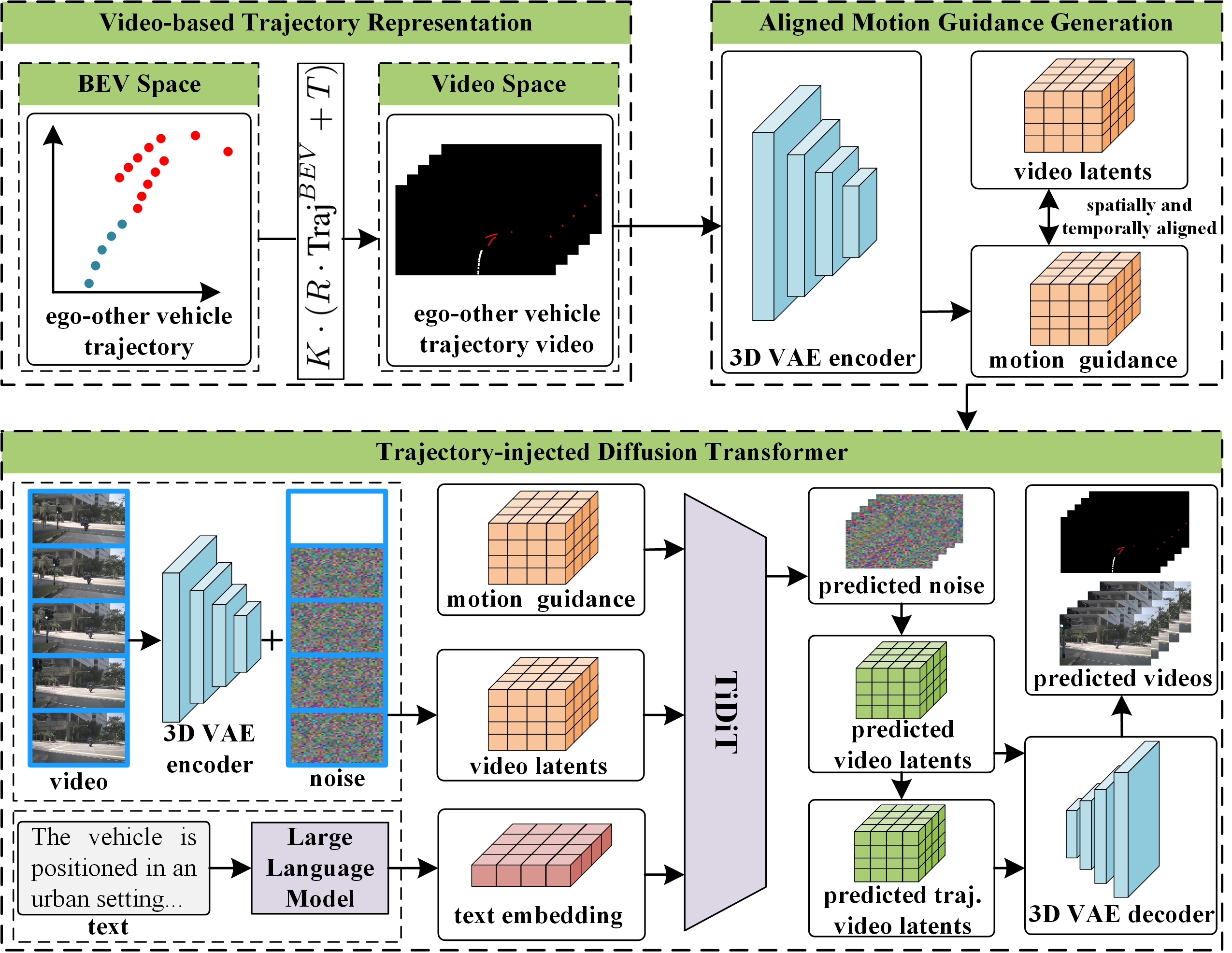}
   \caption{Illustration of the proposed EOT-WM.
   } \label{fig2}
\end{figure*}

\subsection{World Model}

The world model can predict the future states based on historical observations and future actions, which has been extensively applied in simulated games~\cite{ha2018recurrent, hafner2020dream, hafner2021mastering} and indoor embodiment~\cite{koh2021pathdreamer, wang2023dreamwalker, mendonca2023structured}. Recently, the world model has attracted great attention in autonomous driving~\cite{wang2023drivedreamer, lu2024wovogen, wang2024driving, yang2024generalized, gao2024vista, hassan2025gem, agarwal2025cosmos}. WoVoGen~\cite{lu2024wovogen} and Drive-WM~\cite{wang2024driving} are devoted to generating 6-view driving videos based on the given conditions. Drivedreamer~\cite{wang2023drivedreamer}, GenAD~\cite{yang2024generalized}, Vista~\cite{gao2024vista}, and GEM~\cite{hassan2025gem} develop world models to predict future driving scenes via the front-view video. However, most world models mentioned above concentrate on the actions of the ego vehicle and ignore those of the other vehicles, making the interaction with the environment insufficient in these world models. Although GEM~\cite{hassan2025gem} uses future object features and human poses to generate future videos, it is more impractical for autonomous driving systems to obtain them compared with future trajectories. Particularly, Cosmos~\cite{agarwal2025cosmos} is a commercial model comprising over 7 billion parameters, which imposes substantial computational overhead during both training and inference. In this paper, the proposed EOT-WM uses the ego-other vehicle trajectories to represent the motions of vehicles in the scene, and generate more realistic videos with all vehicles controllable.

\section{EOT-WM Framework}

The overall architecture of the proposed EOT-WM framework is illustrated in Fig.~\ref{fig2}. We build EOT-WM based on CogvideoX~\cite{yang2024cogvideox}, and modify it as the world model architecture with injected trajectory condition. Specifically, the proposed EOT-WM framework consists of Video-based Trajectory Representation~(VTR), Aligned Motion Guidance Generation~(AMGG), and Trajectory-injected Diffusion Transformer~(TiDiT), which are described in detail in the following sections. 

\begin{figure}[htbp]
   \centering
   \includegraphics [width=1.0\linewidth] {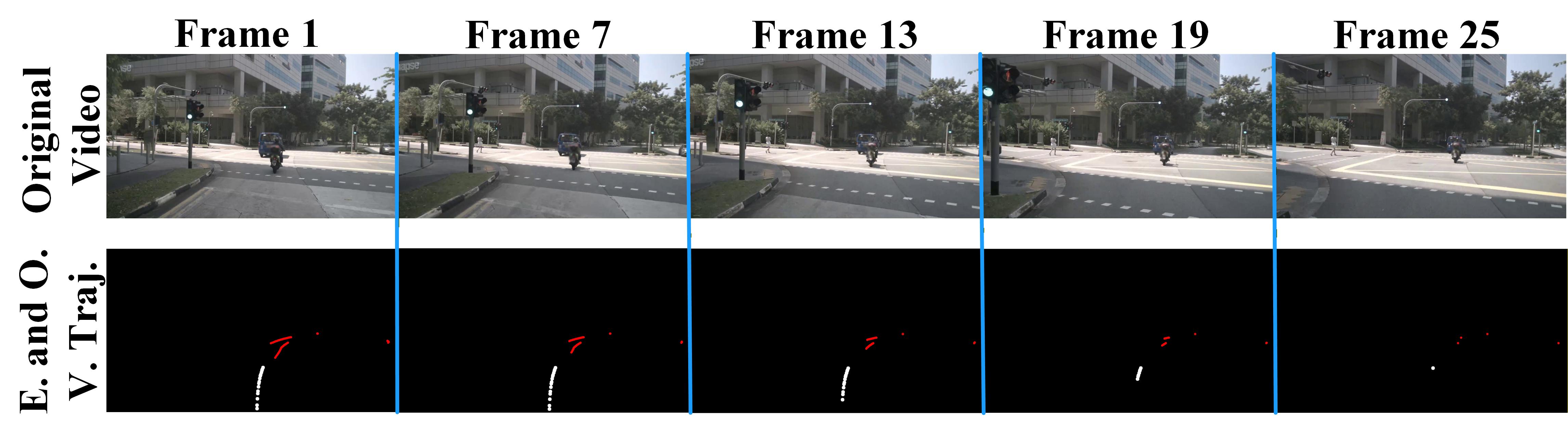}
   \caption{Illustration of the original video, other vehicle trajectory~(O. V. Traj.) and ego vehicle trajectory~(E. V. Traj.) used for the proposed EOT-WM. To be brief, we only visualize the 1st, 7th, 13th, 19th, 25th frames.
   } \label{fig3}
\end{figure}

\subsection{Video-based Trajectory Representation}

The trajectory in the end-to-end autonomous driving system is usually represented as a series of points in the BEV space for planning, which can be formulated as $\text{Traj}^{BEV}=\{(x_{1}, y_{1}), (x_{2}, y_{2}), ..., (x_{T}, y_{T})\}$ and $T$ is the number of points. However, the world model in this paper aims to generate a video that simulates the driving scene based on vehicle trajectories. All elements in the generated video such as vehicles and roads are represented by pixels, leading to the natural mismatch between the points in the BEV space and the pixels in the video. In addition, it is impractical to correspond multiple trajectories in the BEV space to numerous vehicles in the video. Therefore, we propose to represent the trajectory in the form of video to act as the condition. 

Given $T$ trajectory points in the BEV space and $T$ video frames, each trajectory point is projected into the image coordinate of the first frame based on the camera extrinsic and intrinsic parameters. The obtained points are denoted as $\text{Traj}^{I}=\{(x_{1}^{I}, y_{1}^{I}), (x_{2}^{I}, y_{2}^{I}), ..., (x_{T}^{I}, y_{T}^{I})\}$. The projection can be calculated as
\begin{equation}
    \text{Traj}^{I}= K \cdot (R \cdot \text{Traj}^{BEV}+B),
\end{equation}
where $R$ and $B$ are rotation matrix and translation matrix of the camera extrinsic parameters for the first frame, and $K$ is the camera intrinsic parameters. Then, as is shown in Fig.~\ref{fig3}, we plot the ego vehicle trajectory and other vehicle trajectories in separate blank videos to generate trajectory videos. For frame $t$ in the trajectory video, only current and future points $\text{Traj}^{plot}=\{(x_{t}^{I}, y_{t}^{I}), ..., (x_{T}^{I}, y_{T}^{I})\}$ are plotted, indicating the future motion of the vehicle. As a result, the trajectory video contains the motion information of vehicles and corresponds each trajectory to the vehicle in the video.

\subsection{Aligned Motion Guidance Generation}

After obtaining trajectory videos, the trajectory can be learned in unified visual modality with driving scene videos. The world model needs to encode trajectory videos as conditions to guide video generation. Here, we adopt the Spatial-Temporal Variational Auto Encoder~(STVAE) in CogvideoX~\cite{yang2024cogvideox} to encode them. On the one hand, STVAE is capable of extracting high-quality spatial-temporal features of trajectory videos, which are aligned with driving video latents temporally and spatially. On the other, the condition features can share the feature space with the driving video latents since they both use the same STVAE to generate latents, which is easier to be learned jointly. The  vehicle trajectory latents $z_{traj}$ and original video latents $z_{vid}$ for training can be formulated as follows:
\begin{equation}
    z_{traj}=\Psi(V_{traj}),
\end{equation}
\begin{equation}
    z_{vid}=\Psi(V_{ori}),
\end{equation}
where $V_{traj}$, $V_{ori}$ are the ego vehicle trajectory video and original driving video, respectively. $\Psi$ denotes the STVAE in the model. $z_{traj}$ is used to provide motion guidance for video generation. 

\subsection{Trajectory-injected Diffusion Transformer}
Typical Diffusion Transformer in CogvideoX~\cite{yang2024cogvideox} aims to predict the noise $n^{a}$ added to $z_{vid}$ based on the noisy video latent $z_{vid}^{noise}$, which is formulated as
\begin{equation}
    z_{vid}^{noise} = z_{vid}+n^{a}.
\end{equation}
In the inference stage, Denoising Diffusion Probabilistic Model~(DDPM)~\cite{ho2020denoising} can denoise from random sampling noise to video latents for the VAE decoder to generate video. However, the video cannot be generated with specific initial frames in such a manner, which is important for the world model. In addition, the original CogvideoX~\cite{yang2024cogvideox} can only use the text as the condition to control video generation.

Therefore, we design Trajectory-injected Diffusion Transformer~(TiDiT) to predict future driving scenes based on specific initial frames and trajectories. Specifically, to enable the model to generate the video from historical $T_{c}$ frames, we replace the first $T_{c}$ frames in noisy video latents $z_{vid}^{noise}$ with the first $T_{c}$ frames of $z_{vid}$. The process can be realized via a conditioning mask $M=\{t \leq T_{c} | 1 \leq t \leq T\}$, which is formulated as
\begin{equation}
    z = M*z_{vid}+(1-M)*z_{vid}^{noise}.
\end{equation}
To further inject trajectory conditions, frame latents $z$ and motion guidance $z_{traj}$ sharing the feature space are concatenated in channel dimension. As a result, the video and condition latents are fused and aligned spatially and temporally. Then, 3D convolution layers are adopted for learning to obtain the final visual latents $z_{vid}^{f}$, which can be calculated as
\begin{equation}
    z_{vid}^{f} = \text{3d\_conv}(\text{concat}([z, z_{traj}])).
\label{eq8}
\end{equation}

The original Diffusion Transformer in CogvideoX~\cite{yang2024cogvideox} uses the text as the condition to control video generation. To fully reserve the ability of CogvideoX, we utilize the scene caption provided in OmniDrive~\cite{wang2024omnidrive} to train the proposed TiDiT. As a result, visual latents $z_{vid}^{f}$ and text embedding $z_{text}$ extracted from the scene caption are concatenated and fed into several layers of Expert Diffusion Transformer in CogvideoX to predict the noise added to $z_{vid}$ as follows:
\begin{equation}
    n^{p} = \phi(\text{concat}([z_{vid}^{f}, z_{text}])),
\label{eq9}
\end{equation}
where $\phi$ represents the Expert Diffusion Transformer in CogvideoX. Given the added noise $n^{a}$ and the predicted noise $n^{p}$, the diffusion loss $l$ of the proposed model can be formulated as
\begin{equation}
l_{diff}=\frac{{\sum}_{i=1}^{T} (n^{p}_{i}-n^{a}_{i})*M_{i}}{T}.
\end{equation}

In the inference stage, we use the STVAE to extract latents of the initial frames, and replace the corresponding frames of the random sampling noise to obtain $z$. Then, following Eq.~\ref{eq8} and Eq.~\ref{eq9}, the Expert Diffusion Transformer infers the noise for the VAE decoder to generate video.

{\bf Evaluate Trajectory Controllability in Video Latent Space.} To quantitatively and qualitatively evaluate trajectory controllability, we propose a metric based on control latent similarity in the video latent space. Specifically, we first calculate the predicted video latents $z_{vid}^{f}$ according to the prediction of TiDiT, which can be formulated as
\begin{equation}
z_{vid}^{p} = z_{vid}^{noise}-n^{p}.
\end{equation}
Then, we adopt a two-layer MLP to decode the predicted video latents as the predicted trajectory video latents $z_{traj}^{p}$ as follows
\begin{equation}
z_{traj}^{p} = \Omega(z_{vid}^{p}),
\end{equation}
where $\Omega$ represents the MLP. We design a control latent similarity loss $l_{traj}$ to train the MLP, which is formulated as 
\begin{equation}
l_{traj}=\frac{{\sum}_{i=1}^{T} (z_{traj}^{p}-z_{traj})}{T},
\end{equation}
which also serves as the metric to evaluate trajectory controllability. Moreover, the predicted trajectory video latents $z_{traj}^{p}$ can be decoded via STVAE to generate trajectory video for visualization.

\begin{figure*}[ht]
   \centering
   \includegraphics [width=0.7\linewidth] {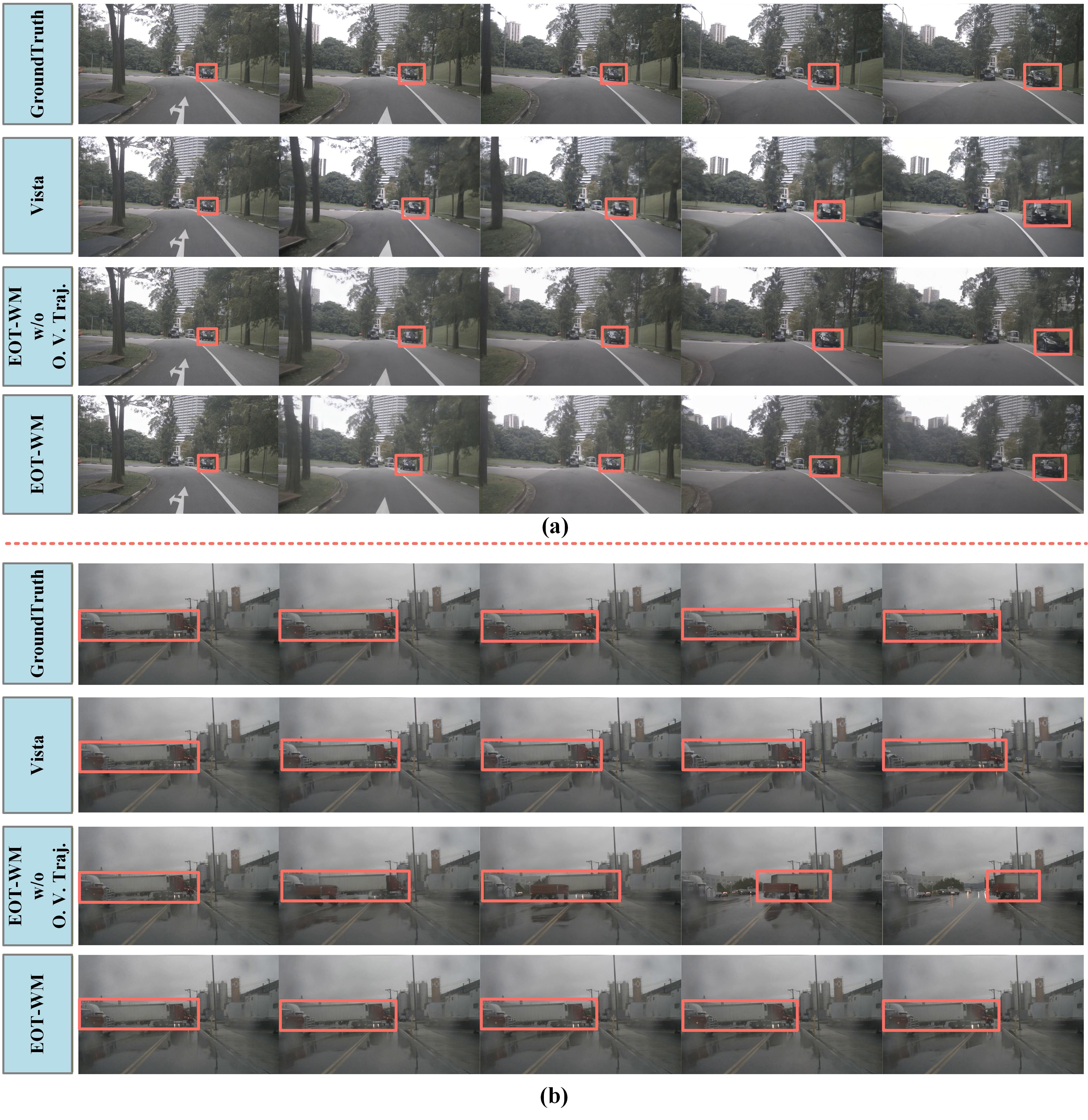}
   \caption{Representative cases for action controllability achieved by the proposed EOT-WM and Vista on the validation set of nuScenes dataset, where EOT-WM w/o O.V. Traj. means the proposed model without learning other vehicle trajectories.} \label{fig4}
\end{figure*}

\section{Experiments}

\subsection{Dataset and Evaluation Metric}
Extensive experiments are conducted on the nuScenes dataset~\cite{caesar2020nuscenes} to evaluate the proposed EOT-WM. Following the setup of Vista~\cite{gao2024vista} in the nuScenes dataset, the training and evaluation set contain 25109 and 5369 videos with 25 frames. The ego vehicle trajectory corresponding to the video can be directly obtained in the annotations. The other vehicle trajectory is only annotated in key frames, and we use interpolation based on the annotated trajectories to complete the full trajectory corresponding to the 25-frame video. We use FID~\cite{heusel2017gans} and FVD~\cite{unterthiner2018towards} to evaluate the quality of generated videos. Moreover, we provide qualitative demonstrations that underscore the advancements of the proposed EOT-WM.

\subsection{Implementation Details}

We build the proposed EOT-WM based on CogvideoX-2B~\cite{yang2024cogvideox}. We use the first video latent as the context for predicting the future frames. The frame rate of the generated video is 10Hz. We use the AdamW~\cite{loshchilov2017decoupled} optimizer with a learning rate of $2\times 10^{-5}$ to train the model. By default, the proposed model is trained at 768×1280 resolution for 60 epochs with 64 NVIDIA A800 GPUs with a total batch size of 128.

\subsection{Comparison with State-of-the-art Methods}

{\bf Quantitative Result.} Comparison results of FID and FVD for video generation achieved by the proposed EOT-WM and competing methods on the validation set of nuScenes dataset are given in Tab.~\ref{tab1}. As can be observed, the state-of-the-art driving world model Vista~\cite{gao2024vista} can generate videos with higher quality and resolution compared with previous works~\cite{santana2016learning, wang2023drivedreamer, lu2024wovogen, wang2024driving, yang2024generalized}. However, Vista~\cite{gao2024vista} cannot control other vehicles in the scene, and using trajectories in BEV space is not sufficiently effective. Therefore, the proposed EOT-WM controlling video generation with ego and other vehicle trajectories in video space achieves 4.8 FID and 40.0 FVD, surpassing Vista by 30\% in FID and 55\% in FVD. Moreover, the proposed EOT-WM can generate videos in 768×1280 resolution, superior to previous world models as well.

\begin{figure*}[ht]
   \centering
   \includegraphics [width=0.72\linewidth] {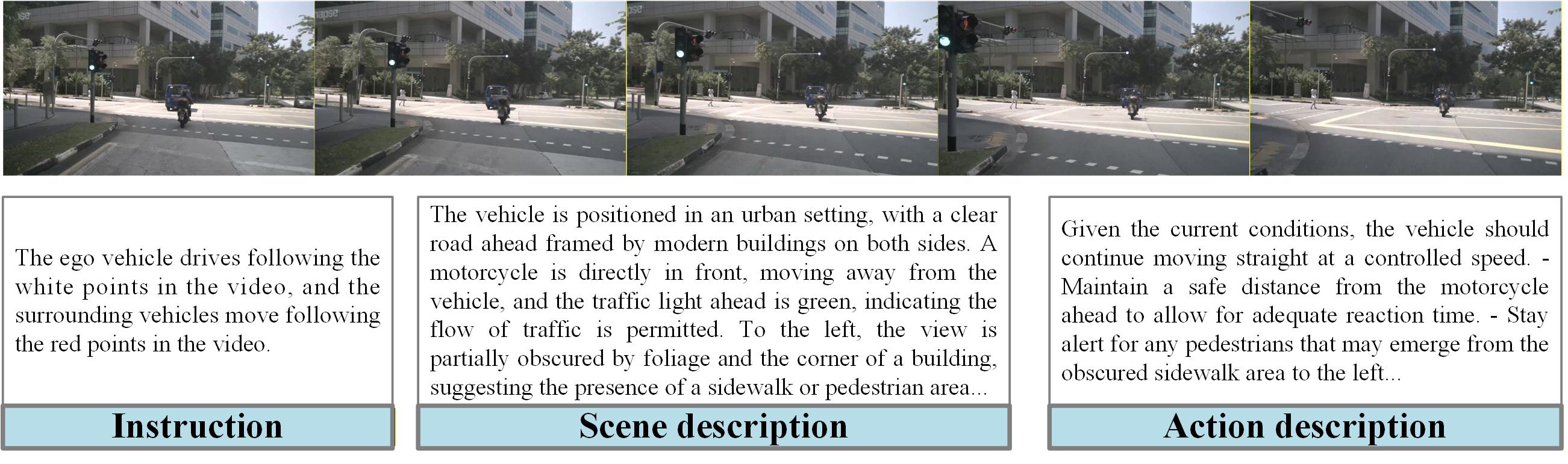}
   \caption{Instance of difference text types used for the proposed EOT-WM, where the scene description and action description are provided in OmniDrive~\cite{wang2024omnidrive}.} \label{fig6}
\end{figure*}
\begin{figure*}[ht]
   \centering
   \includegraphics [width=0.72\linewidth] {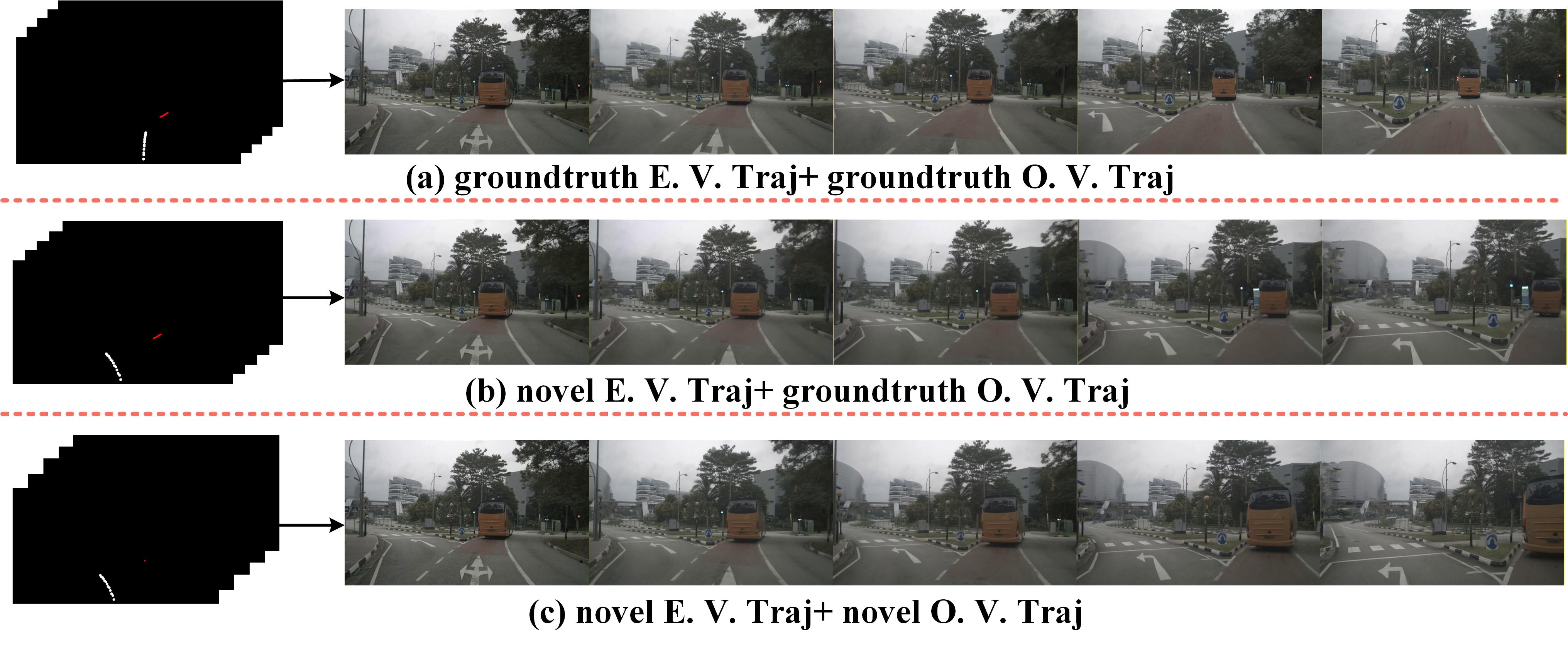}
   \caption{Instances generated with novel trajectories by the proposed EOT-WM.
   } \label{fig7}
\end{figure*}
\begin{figure*}[ht]
   \centering
   \includegraphics [width=0.72\linewidth] {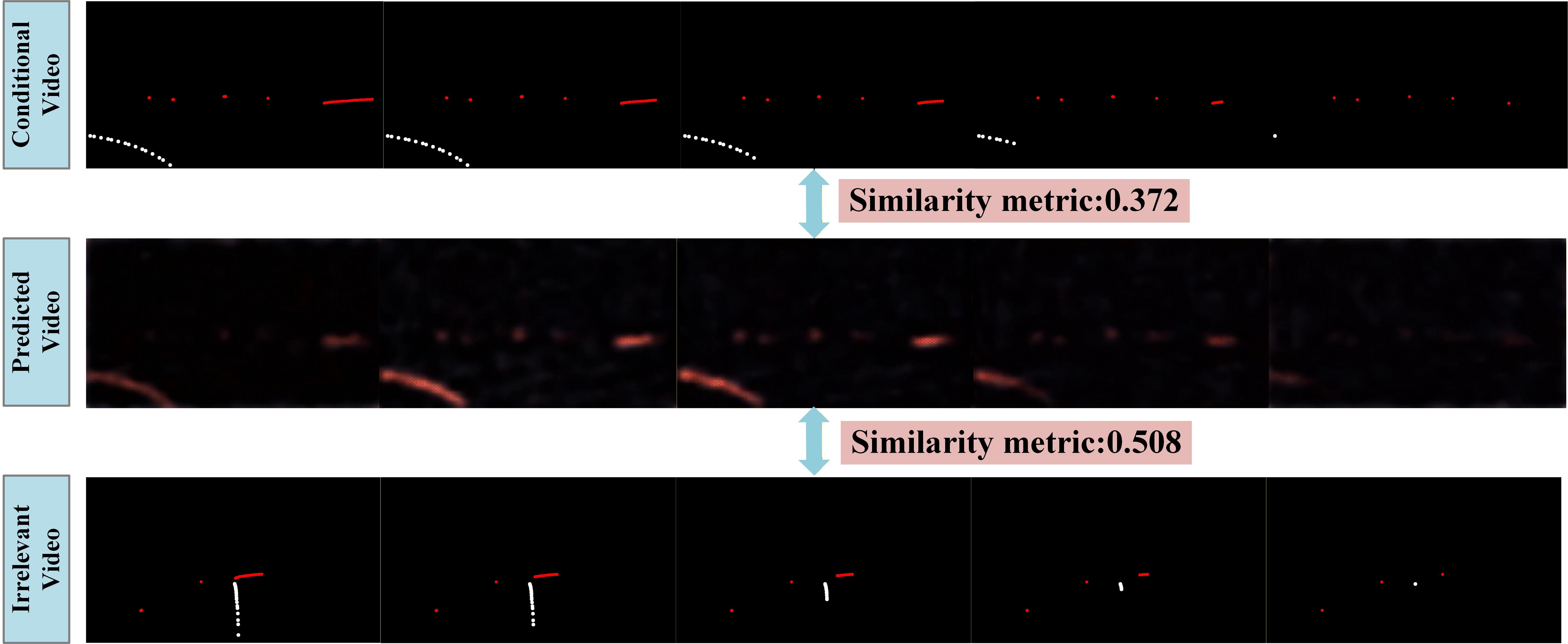}
   \caption{Instance of the proposed metric for evaluating trajectory controllability.} \label{fig8}
\end{figure*}

\begin{table}[htbp]
\renewcommand\arraystretch{1.0}
\setlength{\tabcolsep}{2pt}
\centering
\newsavebox{\tableboxTabOV}
\begin{lrbox}{\tableboxTabOV}
\begin{tabular}{c|c|c|c}
\toprule[1pt]
Methods           & Resolution                  & FID $\downarrow$     & FVD $\downarrow$\\
\hline
DriveGAN~\cite{santana2016learning}    &  256×256 & 73.4          & 502.3      \\
DriveDreamer~\cite{wang2023drivedreamer} & 128×192  & 52.6          & 452.0      \\
WoVoGen~\cite{lu2024wovogen}    &  256×448  & 27.6          & 417.7      \\
Drive-WM~\cite{wang2024driving}   &  192×384  & 15.8          & 122.7      \\
GenAD~\cite{yang2024generalized}  &  256×448  & 15.4         & 184.0      \\
Vista~\cite{gao2024vista}       &  576×1024   & 6.9          & 89.4      \\
\hline
EOT-WM               &    768×1280             & {\bf 4.8}    & {\bf 40.0}    \\
\toprule[1pt]
\end{tabular}
\end{lrbox}
\scalebox{0.95}{\usebox{\tableboxTabOV}}
\caption{Comparison results of FID and FVD for video generation achieved by the proposed EOT-WM and competing methods on the validation set of nuScenes dataset.}
\label{tab1}
\end{table}

\begin{table}[htbp]
\renewcommand\arraystretch{1.0}
\centering
\begin{lrbox}{\tableboxTabOV}
\begin{tabular}{cccc|cc}
\toprule[1pt]
VTR &AMGG &E. V. Traj. & O. V. Traj.& FID $\downarrow$ & FVD $\downarrow$ \\
\hline
- &- &- &- & 61.4 & 372.8 \\
- &\ding{51} &\ding{51} &\ding{51}   & 55.3 & 348.1  \\
\ding{51} &- &\ding{51} &\ding{51}   & 26.1 & 227.2  \\
\ding{51} &\ding{51} &- &\ding{51}  & 49.6 & 312.5  \\
\ding{51} &\ding{51} &\ding{51} &-  & 7.2 & 66.1  \\
\ding{51} &\ding{51} &\ding{51} &\ding{51}   & {\bf 4.8}   & {\bf 40.0}  \\
\toprule[1pt]
\end{tabular}
\end{lrbox}
\scalebox{0.93}{\usebox{\tableboxTabOV}}
\caption{Ablation for key components on the validation set.}
\label{tab2}
\end{table}

{\bf Qualitative Result.} To further demonstrate the superiority in action controllability of the proposed EOT-WM, we provide several representative cases achieved by the proposed EOT-WM compared with Vista~\cite{gao2024vista} in Fig.~\ref{fig4}. As is shown in Fig.~\ref{fig4}~(a), the blank car in the red rectangle stands by the side of the road. Vista and EOT-WM w/o O.V. Traj. do not consider the other vehicle trajectories, and the blank car moves forward in the generated videos. In contrast, the blank car keeps still in the video generated by the proposed EOT-WM. In Fig.~\ref{fig4}~(b), the truck is backing up on the road. However, Vista and EOT-WM w/o O.V. Traj. both generate the truck moving forward in the generated video. As for the proposed EOT-WM, the action of the truck is predicted correctly in the generated video. These cases demonstrate that the proposed EOT-WM can generate more realistic videos with ego and other vehicle trajectories.

\subsection{Ablation Study}

To verify the effectiveness of the proposed modules, the ablation study for key components is conducted on the validation set of nuScenes dataset, and the results are given in Tab.~\ref{tab2}. When all components are not used, the model is close to action-free only with initial frames as the context. As a result, the model obtains poor performance as 61.4 FID and 372.8 FVD. As for {\bf EOT-WM w/o VTR} model, the performance is still poor, since the model cannot learn trajectories in BEV space well and other modules designed for VTR cannot work. For {\bf EOT-WM w/o AMGG} model, we use C3D to extract trajectory video features, and the performance declines compared with the proposed EOT-WM due to the dissimilarity of the feature space. The ego vehicle trajectory is the most important condition for video generation. Therefore, {\bf EOT-WM w/o E. V. Traj.} model cannot generate realistic videos. With other vehicle trajectories to control the other vehicles in the scene, the proposed  {\bf EOT-WM} model can achieve better FID and FVD compared with {\bf EOT-WM w/o O. V. Traj.} model. All above results demonstrate the effectiveness of the proposed modules.

\subsection{Effect of difference text types}

The original CogvideoX~\cite{yang2024cogvideox} is trained to use the text to control video generation, and other advanced video generation models~\cite{guoanimatediff, kong2024hunyuanvideo} are also text-to-video models. When building the world model based on such text-to-video models, it is unadvisable to remove the text element since the capability of the original text-to-video model will be damaged. As can be observed in Tab.~\ref{tab3}, while the proposed EOT-WM does not use the text, FID and FVD achieved decline obviously. Then, we adopt three types of texts to control the video generation model. The performances obtained by these types of texts are similar, demonstrating that the video generation model can be trained to adapt different types of texts. Among them, EOT-WM using the action description achieves the best performance since the caption contains action information. However, considering the capability of generating novel scenes, we choose the scene description for EOT-WM as the action description will conflict with the novel trajectories.

\subsection{Generating novel scenes}

To verify the capability of the proposed EOT-WM to generate novel scenes, several instances generated with novel trajectories are given in Fig.~\ref{fig7}. As can be observed in Fig.~\ref{fig7}~(a), the ego vehicle turns right and the bus moves forward in the video generated with groundtruth ego and other vehicle trajectories in the dataset. While we use the self-produced ego vehicle trajectory that represents turning left in Fig.~\ref{fig7}~(b), the ego vehicle is controlled to turn left in the video generated. Moreover, we give the novel other vehicle trajectory meaning stop in Fig.~\ref{fig7}~(c), the bus in the generated video stands by. Above results demonstrate that the proposed EOT-WM can generate novel scenes according to novel trajectories.

\subsection{Evaluating Trajectory Controllability}

To evaluate trajectory controllability, we propose a metric based on control latent similarity in the video latent space. As is shown in Fig.~\ref{fig8}, we use the conditional trajectory video to generate predicted trajectory video latents in TiDiT. The predicted trajectory video latents can be decoded into predicted trajectory video, which is similar to the conditional trajectory video. Moreover, given another irrelevant trajectory video, the metric calculated between the predicted trajectory video latents and the conditional trajectory video latents is obviously smaller than that calculated between the predicted trajectory video latents and the irrelevant trajectory video latents. These results demonstrate the effectiveness of the proposed latent-based metric.

\begin{table}[htbp]
\renewcommand\arraystretch{1.0}
\centering
\begin{lrbox}{\tableboxTabOV}
\begin{tabular}{c|cc}
\toprule[1pt]
Text Type & FID $\downarrow$ & FVD $\downarrow$ \\
\hline
None & 18.5 & 156.3  \\
Instruction & 5.2 & 42.1  \\
Scene Description & 4.8 & 40.0  \\
Action Description & {\bf 4.1} & {\bf 37.6}  \\
\toprule[1pt]
\end{tabular}
\end{lrbox}
\scalebox{1.0}{\usebox{\tableboxTabOV}}
\caption{Effect of difference texts used in EOT-WM.}
\label{tab3}
\end{table}

\section{Conclusion}

In this work, a novel driving world model named EOT-WM is proposed, which can use ego and other vehicle trajectories to generate realistic videos. Specifically, we propose to represent ego and other vehicle trajectories in the video space instead of the BEV space for learning in the unified visual modality. Then, AMGG is proposed to generate trajectory latents aligned with driving video latents spatially and temporally, which also share feature space. Finally, TiDiT is designed to denoise the noisy video latents more precisely via jointly using the motion guidances based on the ego-other vehicle trajectories. Experiments conducted on the nuScenes dataset demonstrate the superiority of the proposed EOT-WM. In the future work, we will explore the manner to control the other vehicles more precisely and the quantitative metric to evaluate the consistency between the given trajectory and the generated video.

%\bibliography{ref}

\end{document}